\pdfoutput=1

\documentclass[11pt]{article}

\usepackage[]{acl}

\usepackage{times}
\usepackage{latexsym}
\usepackage{array}
\usepackage{CJKutf8}
\usepackage{tcolorbox}
\usepackage{amssymb}
\usepackage{makecell}
\usepackage{latexsym}
\usepackage{diagbox}
\usepackage{enumitem}
\usepackage{amsmath}
\usepackage{subfigure}
\usepackage{multirow}
\usepackage{booktabs}

\def\huggingface{\raisebox{-1.5pt}{\includegraphics[height=1.05em]{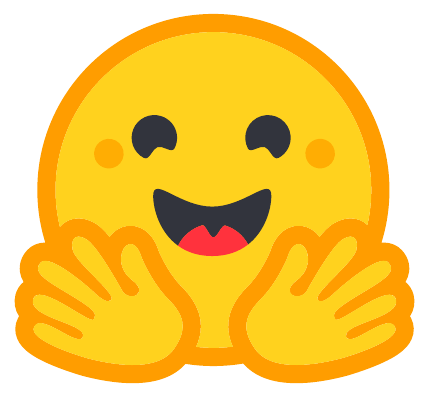}}}
\def\github{\raisebox{-1.5pt}{\includegraphics[height=1.0em]{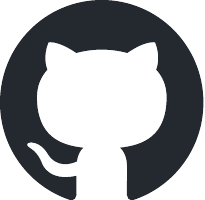}}}

\usepackage[T1]{fontenc}

\usepackage[utf8]{inputenc}

\usepackage{microtype}

\usepackage{graphicx}

\title{ExTrans: Multilingual Deep Reasoning Translation via Exemplar-Enhanced Reinforcement Learning}

\author{Jiaan Wang, \ Fandong Meng\thanks{ \ \ Corresponding author.}, \ Jie Zhou \\
Pattern Recognition Center, WeChat AI, Tencent Inc \\ 
\texttt{\{torchwang,fandongmeng,withtomzhou\}@tencent.com} \\
\github \enspace \href{https://github.com/krystalan/DRT}{ExTrans} \quad \huggingface \enspace \href{https://huggingface.co/Krystalan/ExTrans-7B}{ExTrans-7B} \quad \huggingface \enspace \href{https://huggingface.co/Krystalan/mExTrans-7B}{mExTrans-7B}
}

\begin{document}
\maketitle
\begin{abstract}
In recent years, the emergence of large reasoning models (LRMs), such as OpenAI-o1 and DeepSeek-R1, has shown impressive capabilities in complex problems, \emph{e.g.}, mathematics and coding. Some pioneering studies attempt to bring the success of LRMs in neural machine translation (MT).
They try to build LRMs with deep reasoning MT ability via reinforcement learning (RL).
Despite some progress that has been made, these attempts generally focus on several high-resource languages, \emph{e.g.}, English and Chinese, leaving the performance on other languages unclear.
Besides, the reward modeling methods in previous work do not fully unleash the potential of reinforcement learning in MT.
In this work, we first design a new reward modeling method that compares the translation results of the policy MT model with a strong LRM (\emph{i.e.}, DeepSeek-R1-671B), and quantifies the comparisons to provide rewards.
Experimental results demonstrate the superiority of the reward modeling method. Using Qwen2.5-7B-Instruct as the backbone, the trained model achieves the new state-of-the-art performance in literary translation, and outperforms strong LRMs including OpenAI-o1 and DeepSeeK-R1.
Furthermore, we extend our method to the multilingual settings with 11 languages.
With a carefully designed lightweight reward modeling in RL, we can simply transfer the strong MT ability from a single direction into multiple (\emph{i.e.}, 90) translation directions and achieve impressive multilingual MT performance.

\end{abstract}

\section{Introduction}

Recently, Large Reasoning Models (LRMs), \emph{e.g.}, OpenAI-o1~\cite{openai_o1_2024} and DeepSeek-R1~\cite{guo2025deepseek}, have shown promising performance in various complex tasks like mathematics, coding, question-answering and search engine~\cite{chen2025towards,li2025system,zhang2024o1,guan2025deeprag,jin2025search,li2025search}.
With the help of the long chain-of-thoughts (CoT), LRMs can make deep thoughts and analyses for a given task, and thus provide thoughtfully verified and explainable results.

In view of the strong ability of LRMs, some researchers attempt to study the long CoT in neural machine translation (MT).
\citet{zhao2024marco} propose Marco-o1, which studies the long CoT in open-ended text generation. They use several examples to briefly show the effectiveness of LRMs in MT with slang or colloquial expressions.
\citet{chen2025evaluating} and \citet{liu2025new} conduct empirical studies on how LRMs perform on MT, and further demonstrate the potential of LRMs in MT.
\citet{wang2024drt} construct MetaphorTrans, a MT dataset on the literary domain, which contains English literary sentences with synthesized long CoT for translation, along with their corresponding Chinese translations.
Based on MetaphorTrans, they train DRT LRMs via supervised fine-tuning (SFT), and demonstrate their effectiveness in English-to-Chinese literary MT.

More recently, after the emergence of DeepSeek-R1~\cite{guo2025deepseek}, the significance of reinforcement learning (RL) in LRMs has become increasingly evident.
Several studies try to leverage RL to enhance the MT ability of LRMs.
R1-T1~\cite{he2025r1} and MT-R1-Zero~\cite{feng2025mt} utilize Comet~\cite{rei-etal-2020-comet}, CometKiwi~\cite{rei-etal-2022-cometkiwi} or BLEU~\cite{papineni-etal-2002-bleu} scores as the rewards to optimize LRMs via RL.
DeepTrans~\cite{wang2025deep} leverages DeepSeek-v3~\cite{liu2024deepseek} as the reward model to provide reference-free evaluation scores in RL training.
However, these reward modeling methods (especially Comet and CometKiwi) are not suitable for all domains~\cite{karpinska-iyyer-2023-large,wang2024drt}, and might result in sub-optimal solutions. Besides, existing work typically focuses on high-resource languages, \emph{e.g.}, English and Chinese, and neglects the transferability of LRMs across different languages.
Therefore, we argue that the improvement of MT ability brought by RL is still under-explored, especially in multilingual MT.

In this paper, our goals are twofold: (1) \emph{\textbf{Unleashing the potential of RL in MT LRMs.}}
The essence of the reward modeling in the MT task is to use different signals to quantify the translation quality.
R1-T1 and MT-R1-Zero only use traditional metrics (Comet, CometKiwi or BLEU) as the reward signals, ignoring the potential ability of large language models (LLMs).
DeepTrans leverages the strong ability of LLM-as-a-judge, and it uses DeepSeek-v3 to quantify the translation quality in a reference-free manner.
In addition to LLM-as-a-judge, we want to further explore the strong ability of \emph{LLM-as-an-exemplar} in the reward modeling.
The ability of LLM-as-an-exemplar is well adopted in knowledge distillation~\cite{xu2024survey}, where an advanced LLM serves as an exemplar (or a teacher) to generate training data for another small model.
In MT, we use an advanced LLM (\emph{i.e.}, DeepSeek-R1) to generate exemplar translations for source sentences.
During RL training, the translations of the policy LRM are compared with the exemplar translations to provide reward signals. For example, if the policy LRM provides better translations than exemplar translations, higher rewards should be assigned to encourage it.
The comparisons can also be made by an advanced LLM (\emph{e.g.}, DeepSeek-v3).
In this manner, both the ability of LLM-as-a-judge and LLM-as-an-exemplar can be adopted in the reward modeling.
Intuitively, a strong exemplar can establish lower bounds for the policy model, potentially guiding it towards enhanced performance.
(2) \emph{\textbf{Transferring the success of MT LRMs to the multilingual settings.}}
Previous work~\cite{wang2024drt,feng2025mt,wang2025deep} generally focuses on high-resource languages, and we want to transfer the successes from a single to multiple translation directions.
However, the ability of LLM-as-a-judge and LLM-as-an-exemplar is not ensured in low-resource languages~\cite{zhu-etal-2024-multilingual,son2024mm}.
Thus, we design a lightweight reward modeling method that only uses LLMs' strong ability in high-resource MT reward modeling (\emph{i.e.}, English-to-Chinese). When translating in other directions during RL training, we focus solely on verifying the correctness of the generation format and the target language to provide rewards.
In this way, we avoid the potential inaccuracies of LLMs' ability as the reward model in low-resource languages.
Besides, we can effectively manage the training costs since the reward modeling in other directions can be achieved only via existing lightweight toolkits, \emph{i.e.}, regular expressions and language detection.

Based on the motivation in our goal (1), we first train our \textbf{Ex}emplar-enhanced \textbf{Trans}lation (abbr., ExTrans-7B) model via RL (using Qwen2.5-7B as the backbone) in English-to-Chinese literary translation, following \citet{wang2024drt,wang2025deep}.
The experimental results demonstrate the superiority of our ExTrans-7B.
With the help of the exemplar (DeepSeek-R1), ExTrans-7B achieves the new state-of-the-art performance, and it outperforms strong LRMs (including OpenAI-o1 and DeepSeek-R1) in terms of both automatic metrics and GPT-4o evaluation.
Ablation studies also verify the rationality of our designation.
Furthermore, we extend ExTrans-7B from English-to-Chinese to the multilingual settings with 11 languages (90 translation directions).
The extension uses the lightweight reward modeling method in our goal (2), and we name the extended LRM as mExTrans-7B.
Experimental results on the multilingual scenes show the surprising improvement brought by the lightweight method.
Without quantifying the translation quality in other translation directions, the mExTrans-7B can effectively transfer the success from English-to-Chinese to other directions, and enhance the model performance by a large margin.
mExTrans-7B achieves competitive multilingual MT performances compared with QwQ-32B.

Our main contributions are concluded as follows:
\begin{itemize}[leftmargin=*,topsep=0pt]
\setlength{\itemsep}{0pt}
\setlength{\parsep}{0pt}
\setlength{\parskip}{0pt}
\item We propose a new reward modeling method that employs a strong LRM as an exemplar, and compares the policy model with the exemplar to provide reward signals. In this way, both the strong abilities of LLM-as-a-judge and LLM-as-an-exemplar are adopted in reward modeling. Based on the method, we train our ExTrans-7B.
\item We propose a lightweight method to extend ExTrans-7B from a single translation direction to the multilingual settings (mExTrans-7B). Without quantifying the translation quality in other directions, the model can also generalize the MT ability across different languages.
\item Extensive experiments demonstrate the superiority of ExTrans-7B and mExTrans-7B. ExTrans-7B achieves the state-of-the-art performance in English-to-Chinese, while mExTrans-7B shows the effectiveness in multilingual settings.
Both models outperform baselines in terms of automatic metrics and GPT-4o evaluation.

\end{itemize}

\begin{figure*}[t]
\centerline{\includegraphics[width=0.98\textwidth]{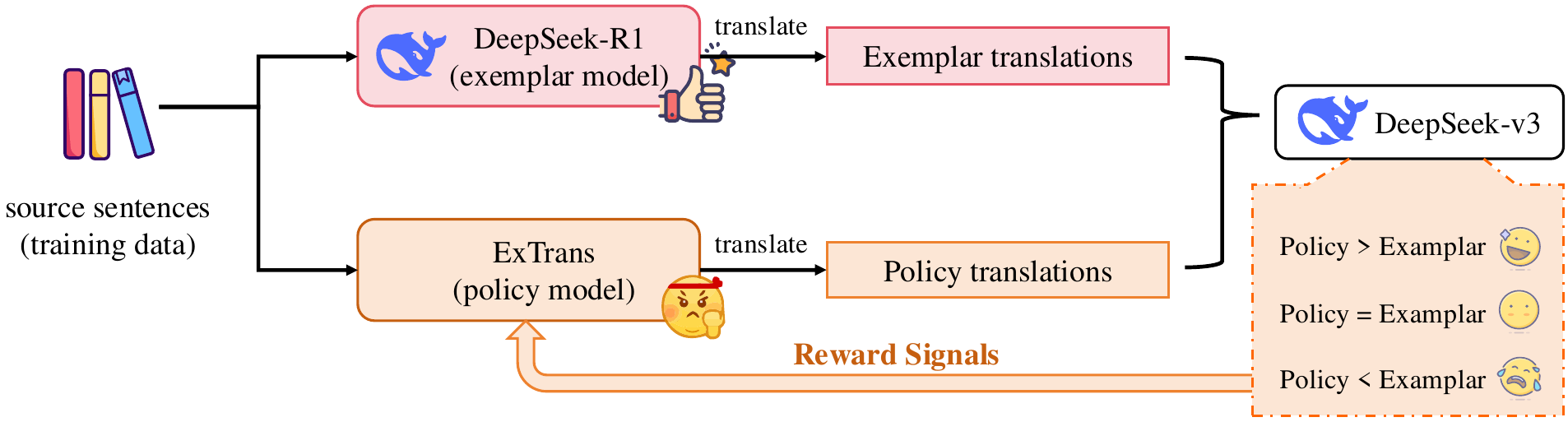}}
\caption{The illustration of the exemplar-enhanced translation reward.}
\label{fig:trans_reward_overall}
\end{figure*}

\section{Methodology}

In this section, we first introduce the proposed reward modeling method, which quantify the translation quality by comparing the translations of the policy model with an exemplar LRM (\S~\ref{subsec:2.1}).
Then, we discuss how to extend the RL training into multilingual settings via lightweight adaptations (\S~\ref{subsec:2.3}).
Finally, we provide the training details of ExTrans-7B and mExTrans-7B (\S~\ref{subsec:2.3}).

\subsection{Reward Modeling}
\label{subsec:2.1}

Given a source sentence $s$, ExTrans first thinks about how to translate $s$ with a long CoT (defined as $t_{h}$) and then provides the corresponding translation (defined as $t_{r}$). The generation format can be defined as ``\texttt{<think>} $t_{h}$ \texttt{</think>} $t_r$'', where ``\texttt{<think>}'' and ``\texttt{</think>}'' are two special tokens to indicate the boundary of $t_h$.

\vspace{0.5ex}
\noindent \textbf{Format and Thought Reward.}
We first use a regular expression to verify the correctness of the generation format, and define the format reward as:
\begin{equation}
  r_{\text{format}} = 
\begin{cases} 
1 & \text{if format is correct} \\
0 & \text{if format is incorrect}
\end{cases}
\end{equation}
The format reward is a basic and commonly-used reward in RL, designed to ensure that the policy model produces outputs in the correct format.

Then, we follow \citet{wang2025deep}, and employ DeepSeek-v3 (671B)~\cite{liu2024deepseek} to assess the quality of the thought process (\emph{i.e.}, $t_h$).
With a carefully designed prompt\footnote{The assessment prompt is shown in Appendix~\ref{appendix:thought_reward_prompt}.}, DeepSeek-v3 could classify $t_h$ into three categories: thought with no analysis, slight analysis or detailed analysis.
Therefore, we can define the thought reward as:
\begin{equation}
\small 
  r_{\text{thought}} = 
\begin{cases} 
3 & \text{if} \operatorname{v3}^{\text{th}}(s, t_h) = \text{detailed analysis} \\
2 & \text{if} \operatorname{v3}^{\text{th}}(s, t_h) = \text{slight analysis} \\
1 & \text{if} \operatorname{v3}^{\text{th}}(s, t_h) = \text{no analysis} \\
\end{cases}
\end{equation}
where $\operatorname{v3}^{\text{th}}$ denotes using DeepSeek-v3 as the thought reward judger.
As pointed out by \citet{wang2025deep}, $r_{\text{thought}}$ can encourage the policy model to generate meaningful thought content, which is important to improve the quality of $t_r$.

\vspace{0.5ex}
\noindent \textbf{Exemplar-Enhanced Translation Reward.}
To leverage the strong ability of LLM-as-an-exemplar, we propose a new method to quantify the translation quality of $t_r$.
As shown in Figure~\ref{fig:trans_reward_overall}, we use an advanced LRM, \emph{i.e.}, DeepSeek-R1 (671B)~\cite{guo2025deepseek}, as an exemplar.
For source sentence $s$, we also use DeepSeek-R1 to generate its translation, named exemplar translation (denoted as $t^\text{e}_r$).
Then, we use DeepSeek-v3 to compare the quality between $t_r$ and $t^\text{e}_r$ using the following prompt:

\begin{tcolorbox}
\fontsize{10pt}{11pt}\selectfont

You are required to evaluate the quality of two translations of a given text from \texttt{[src lang]} to \texttt{[trg lang]}.

The given text is: \{$s$\}\\

For the text, there are two translations:\\
- Translation 1: \{$t^{\text{e}}_r$\}\\
- Translation 2: \{$t_r$\}\\

Please carefully compare the two translations, and determine which of the following situations the two translations belong to: \\

- Situation 1: Translation 1 is significantly better than Translation 2\\
- Situation 2: Translation 1 is slightly better than Translation 2.\\
- Situation 3: The quality of both translations is similar.\\
- Situation 4: Translation 2 is slightly better than Translation 1.\\
- Situation 5: Translation 2 is significantly better than Translation 1.\\

Your assessment should be based on factors such as accuracy, fluency, and overall readability.
\end{tcolorbox}
\noindent where ``\texttt{[src lang]}'' and ``\texttt{[trg lang]}'' denote the source and the target languages. respectively. Next, we can define the exemplar-enhanced translation reward value as:
\begin{equation}
\small 
r_{\text{trans}} = i \ \  \text{if} \operatorname{v3}^{\text{tr}}(s, t_r, t^{\text{e}}_r) = \text{Situation}\ i\ (i \in \{1,2,3,4,5\}) 
\end{equation}
where $\operatorname{v3}^{\text{tr}}$ denotes using DeepSeek-v3 as the translation reward judger.
In this manner, the strong exemplar translation $t^{\text{e}}_r$ can establish lower bounds for the policy model, potentially guiding it towards enhanced performance.
Besides, both the abilities of LLM-as-a-judge and LLM-as-an-exemplar are adopted in our reward modeling.

\vspace{0.5ex}
\noindent \textbf{CometKiwi Reward.}
In addition to $r_\text{trans}$, we also employ CometKiwi~\cite{rei-etal-2022-cometkiwi} as an auxiliary reward to quantify the translation quality:
\begin{equation}
r_{\text{cometk}}(s, t_r) \in [0,1]
\end{equation}

\vspace{0.5ex}
\noindent \textbf{Overall Reward.}
The overall reward is the combination of the above rewards:
\begin{equation}\label{eq:reward_in_single_direction}
\small
  r_{\text{all}} = 
\begin{cases} 
0 & \text{if  } r_{\text{format}} = 0 \\
r_{\text{trans}} + \alpha \times r_{\text{thought}} + \beta \times r_{\text{cometk}} & \text{if  } r_{\text{format}} \neq 0
\end{cases}
\end{equation}
where $\alpha$ and $\beta$ are trade-off coefficients.

\subsection{Multilingual Reward Modeling}
\label{subsec:2.2}

We extend the reward modeling from a single translation direction to multilingual settings.
A straightforward way is to directly use $r_{\text{all}}$ (c.f. Eq.~\ref{eq:reward_in_single_direction}) in the multilingual settings.
However, it suffers from the following issues:
(\romannumeral1) the ability of LLM-as-a-judge and LLM-as-an-exemplar is not ensured in low-resource languages~\cite{zhu-etal-2024-multilingual,son2024mm}.
The effectiveness of utilizing DeeSeek-R1 for generating exemplar translations (\emph{i.e.}, $t^{\text{e}}_r$) or DeepSeek-v3 for providing rewards (\emph{i.e.}, $r_\text{thought}$ and $r_\text{trans}$) in languages other than English and Chinese remains uncertain.
As pointed out by \citet{deepseekv3mteval}, the correlation between DeepSeek-v3/R1 and human judgments in MT evaluation is significantly stronger for Chinese-English translations compared to other directions.
(\romannumeral2) It still needs high computational costs to obtain exemplar translations and rewards in various languages, neglecting the exploration of the internal transferability of MT capabilities across different languages.

Therefore, we design a lightweight method to generalize the MT capabilities to multilingual translation. Specifically, we choose English-to-Chinese as a representative high-resource direction.
During RL training, only if the policy model performs English-to-Chinese translation, we employ $r_\text{all}$ as the rewards to update the model.
For other directions, we solely verify the correctness of the generation format and the target language to provide rewards, which can be trivially achieved via regular expressions and language detection toolkits:
\begin{equation}\label{eq:reward_in_multilingual_direction}
\small
  r_{\text{generalize}} = 
\begin{cases} 
0 & \text{if  } r_{\text{format}} = 0 \\
0 & \text{elif  } \operatorname{Lang}(t_r) \neq \text{target language} \\
1 & \text{else}
\end{cases}
\end{equation}
where $\operatorname{Lang}(\cdot)$ denotes the language detection toolkit, which identifies the language of the model translations (\emph{i.e.}, $t_r$).
In this manner, we can avoid quantifying translation quality in most directions, allowing the policy model to transfer MT capabilities across languages.

\subsection{Training Details}
\label{subsec:2.3}

Following \citet{wang2025deep,feng2025mt}, we use Qwen2.5-7B-Instruct~\cite{yang2024qwen2} as the backbone to train our ExTrans-7B and mExTrans-7B.
The training process includes two stages: cold start SFT and RL training.

\vspace{0.5ex}
\noindent \textbf{Cold Start SFT.}
We first apply SFT using a cold-start dataset that encourages a ``think-then-translate'' pattern.
In detail, we randomly select \emph{general-domain} source sentences from WMT24\footnote{\url{https://www2.statmt.org/wmt24/index.html}}, and use DeepSeek-R1 to generate seed translation samples (with long CoT) in the general domain.
There are 4K English-to-Chinese samples used in the cold start SFT stage of ExTrans-7B.
For the multilingual settings, we focus on 11 languages: English (abbr. En), Chinese (Zh), Arabic (Ar), Czech (Cs), German (De), Spanish (Es), French (Fr), Italian (It), Japanese (Ja), Russian (Ru) and Korean (Ko), resulting in 110 translation directions.
Therefore, in addition to 4K English-to-Chinese samples, we further collect 8K samples for the other 109 directions. Consequently, there are 12K samples used to cold start SFT mExTrans-7B.

\vspace{0.5ex}
\noindent \textbf{RL Training.}
Following~\citet{guo2025deepseek,wang2025deep,feng2025mt}, we use GRPO algorithm~\cite{shao2024deepseekmath} in the RL training stage.
For the policy model $\pi$, given a source sentence $s$, it can roll out a number of generations $\{g_1, g_2, ..., g_n\}$.
GRPO optimizes the policy model $\pi^*$ by maximizing the following objective:
\begin{equation}
\small
\frac{1}{n}\sum_{1}^{n}(\operatorname{min}(\nabla_{\pi}A_i, \operatorname{clip}(\nabla_{\pi}, 1-\epsilon, 1 + \epsilon)A_i) -\beta \mathcal{D}
\end{equation}
\begin{equation}
\small
\nabla_{\pi} = \frac{\pi^*(g_i|s)}{\pi(g_i|s)}
\end{equation}
where $\epsilon$ and $\beta$ are hyperparameters.
$\mathcal{D}$ indicates the KL divergence between the policy model $\pi^*$ and the reference model.
$A_i$ indicates the advantage that is calculated as follows:
\begin{equation}
\small
A_i = \frac{r^{i} - {\operatorname{mean}(\{r^1, r^2, \cdots, r^n\})}}{{\operatorname{std}(\{r^1, r^2, \cdots, r^n\})}}.
\end{equation}
where $r^i$ denotes the reward of $g_i$.
When training ExTrans-7B, we use $r_\text{all}$ (c.f. Eq.~\ref{eq:reward_in_single_direction}) as the reward signals. For training mExTrans-7B, both $r_\text{all}$ and $r_\text{generalize}$ (c.f. Eq.~\ref{eq:reward_in_multilingual_direction}) are leveraged.

\section{Experiments}

\subsection{Experimental Setups}

\noindent \textbf{Training Data in RL.}
Following \citet{wang2024drt,wang2025deep}, we focus on the literary MT, where the translation generally requires cultural background.
For the RL training of ExTrans-7B, we use MetaphorTrans~\cite{wang2024drt}, which contains 19K training English-to-Chinese samples.
We only use the source sentences during RL training, which are selected from English literary books, and generally contain metaphors or similes.
Only English-to-Chinese translation performs in ExTrans-7B training.
For RL training of mExTrans-7B, we focus on 11 languages. In addition to the source English sentences from MetaphorTrans, we also use Par3 data~\cite{thai-etal-2022-exploring} which involves literary sentences in Zh, Cs, De, Es, Fr, It, Ja and Ru.
Thus, the above 8 languages along with English can serve as source languages, each of which could be paired with 10 other languages (\emph{i.e.}, a total of 11 languages minus the source language itself) for translation.
As a result, there are 90 (9$\times$10) directions used in the RL training of mExTrans-7B.
To explore the generalizability across different languages, in addition to 19K English-to-Chinese samples, we only collect 50 samples for the other 89 directions, resulting in \textasciitilde23.5K samples in total.

\vspace{0.5ex}
\noindent \textbf{Evaluation Data.}
We also leverage the test set of MetaphorTrans as our evaluation data, which contains 1K English literary sentences.
Besides, we follow \citet{wang2025deep}, and employ two literature books (in English) as the evaluation data:
(1) \emph{The Essential O. Henry Collection} (by O. Henry) and (2) \emph{Orbital} (by Samantha Harvey).
Both books are rich in literary nuance, making them challenging for even human translators.
When evaluating ExTrans-7B, the above English sentences are used to perform English-to-Chinese translation.
When evaluating mExTrans-7B, English-to-X (X=Zh, Ar, Cs, De, Es, Fr, It, Ja, Ru or Ko) translations are conducted.
To further evaluate the multilingual MT performance, we also randomly select 500 Russian literary sentences from Par3, and conduct Russian-to-X translations.

\vspace{0.5ex}
\noindent \textbf{Evaluation Metrics.}
Since (1) the references in MetaphorTrans are synthesized via LLMs without human annotations; and (2) the references of \emph{The Essential O. Henry Collection} and \emph{Orbital} are missing, we adopt reference-free metrics in our experiments.
Following \citet{wang2025deep,feng2025mt}, we use \emph{CometKiwi}~\cite{rei-etal-2022-cometkiwi} to evaluate the model translations.
Moreover, following \citet{wang2025deep}, we use three evaluators implemented using GPT-4o in reference-free manners, which we refer to as \emph{GRF}, \emph{GEA5} and \emph{GEA100}, respectively.
Among them, \emph{GRF} assesses model performance from a general perspective, while \emph{GEA5} and \emph{GEA100} evaluate it from a literary perspective.
For the details of the evaluation prompts, please refer to Appendix~\ref{appendix:evaluation_prompt}.

\vspace{0.5ex}
\noindent \textbf{Implementation Details}.
For the implementation details of cold-start SFT, RL training, evaluation toolkits, and evaluation hyperparameters, please refer to Appendix~\ref{appendix:implementation_details}.

\begin{table*}[t]
\centering
\resizebox{0.98\textwidth}{!}
{
\begin{tabular}{lcccccccccccc}
\toprule[1pt]
\multicolumn{1}{c}{\multirow{2}{*}{Model}}      & \multicolumn{4}{c}{MetaphorTrans}                                & \multicolumn{4}{c}{\emph{O. Henry}}                                     & \multicolumn{4}{c}{\emph{Orbital}}                                      \\
\cmidrule(r){2-5} \cmidrule(r){6-9} \cmidrule(r){10-13} & GRF            & GEA5          & GEA100         & CometKiwi      & GRF            & GEA5          & GEA100         & CometKiwi      & GRF            & GEA5          & GEA100         & CometKiwi      \\ \midrule[1pt]
Marco-o1-7B               & 82.41 & 3.57 & 64.24  & 71.62     & 83.12 & 3.71 & 63.11  & 76.00        & 81.84 & 3.89 & 67.64  & 75.38     \\
Qwen2.5-7B-Instruct       & 81.53 & 3.62 & 66.21  & 70.36     & 85.26 & 3.83 & 66.50   & 76.18     & 83.38 & 4.00    & 70.10   & 76.46     \\
Llama-3.1-8B-Instruct     & 79.25 & 3.31 & 59.58  & 70.14     & 79.73 & 3.43 & 57.17  & 74.35     & 79.92 & 3.54 & 59.90   & 75.09     \\
DeepSeek-Qwen-7B          & 65.16 & 2.67 & 43.66  & 63.49     & 68.97 & 2.86 & 45.64  & 70.67     & 71.28 & 3.16 & 51.91  & 72.43     \\
DeepSeek-Llama-8B         & 76.31 & 3.24 & 56.89  & 67.13     & 78.17 & 3.39 & 56.14  & 73.39     & 78.91 & 3.64 & 59.75  & 74.47     \\
Qwen3-8B (w/o CoT) & 85.96 & 4.22 & 72.71 &  72.96 & 89.15 & 4.13 & 75.81 & 77.95 & 85.76 & 4.29 & 78.06 & 78.26   \\
Qwen3-8B (w/ CoT)  & 87.02 & 4.22 & 74.00 &  73.13 & 88.96 & 4.10 & 76.44 & 77.90 &  84.73 & \underline{4.32}  & 77.66 & 78.01  \\
Qwen2.5-7B-MT$^{\diamondsuit}$             & 85.06 & 3.93 & 72.29  & 71.03     & 86.84 & 4.05 & 71.05  & 77.29     & 85.46 & 4.12 & 70.55  & 76.32     \\
Llama-3.1-8B-MT$^{\diamondsuit}$           & 84.10 & 3.88 & 69.33  & 70.25     & 85.04 & 3.87 & 66.60   & 76.14     & 80.37 & 3.87 & 64.38  & 75.11     \\
DRT-7B$^{\diamondsuit}$                    & 85.57 & 4.05 & 75.05  & 71.78     & 86.36 & 3.96 & 69.51  & 76.12     & 81.69 & 3.84 & 65.56  & 69.95     \\
DRT-8B$^{\diamondsuit}$                    & 84.49 & 3.91 & 69.65  & 70.85     & 83.61 & 3.75 & 64.76  & 73.89     & 79.14 & 3.65 & 61.36  & 66.36     \\ 
DeepTrans-7B$^{\diamondsuit}$              & \underline{88.84} & 4.21 & 75.38  & 71.82     & 87.95 & \underline{4.22} & 76.92  & 77.04     & \textbf{87.95} & 4.22 & 76.92  & 76.65     \\ \midrule[1pt]
Qwen2.5-14B-Instruct      & 84.74 & 3.87 & 70.86  & 72.01     & 86.83 & 3.98 & 70.53  & 77.04     & 84.39 & 4.09 & 71.65  & 76.55     \\
DeepSeek-Qwen-14B         & 83.92 & 3.81 & 70.64  & 71.01     & 83.27 & 3.82 & 64.79  & 75.22     & 82.30  & 4.01 & 69.10   & 76.28     \\
Qwen3-14B (w/o CoT) & 83.48 & 4.09 & 70.74 &  71.07 &  88.78  & 4.14 & 77.15 & 77.85 & 84.50  & 4.30 & 79.26 & \underline{78.55}    \\
Qwen3-14B (w/ CoT)  & 88.27 & 4.29 & 74.44 &  \underline{73.84} & 89.74 & 4.16 & 77.38 & 77.32 & 85.91 & 4.29 & 79.01 & 77.98   \\
Qwen2.5-14B-MT$^{\diamondsuit}$            & 85.66 & 4.02 & 74.53  & 72.08     & 87.27 & 4.05 & 73.06  & 77.54     & 85.55 & 4.14 & 75.84  & 77.40      \\
DRT-14B$^{\diamondsuit}$                   & 87.19 & 4.13 & 77.41  & 72.11     & 87.38 & 4.00    & 72.59  & 76.70      & 82.19 & 3.98 & 69.36  & 70.99     \\ \midrule[1pt]
DeepSeek-Qwen-32B         & 84.78 & 3.87 & 71.88  & 71.93     & 87.03 & 4.03 & 70.81  & 76.75     & 85.36 & 4.16 & 73.62  & 77.80      \\
QwQ-32B-preview           & 86.31 & 4.00  & 75.50   & 71.48     & 87.61 & 4.03 & 70.79  & 76.86     & 84.79 & 4.04 & 71.03  & 76.17     \\
QwQ-32B                   & 88.06 & 4.09 & 74.38  & 72.88     & 88.02 & 4.21 & 76.36  & 77.71     & \underline{87.83} & 4.15 & 76.55  & 77.55     \\
Qwen3-32B (w/o CoT) & 84.08 & 4.15 & 72.44 &  71.18 & \underline{89.87} & 4.17 & \underline{78.51} & 77.80 & 86.33 & 4.25 & 79.59 & 78.39    \\
Qwen3-32B (w/ CoT)  & 88.29 & \underline{4.33}  & 76.29 &  73.37 & 89.76 & 4.17 & 77.88 & 77.23 & 86.54 & 4.26 & 79.71 & 78.05   \\
GPT-4o                    & 85.57 & 3.86 & 71.88  & 73.01     & 88.30  & 4.00    & 71.06  & 76.74     & 85.91 & 4.17 & 73.54  & 77.67     \\
o1-preview                & 87.11 & 4.06 & \underline{78.01}  & 73.70      & 89.73 & 4.14 & 76.17  & \textbf{78.41}     & 86.85 & 4.26 & 76.80   & \textbf{78.86}     \\
DeepSeek-R1               & 84.29 & 4.02 & 73.78  & 68.33     & 89.79 & 4.17 & 77.03  & 77.01     & 87.37 & 4.27 & \underline{80.06}  & 76.17     \\ \midrule[1pt]
ExTrans-7B (cold start)   & 85.06 & 3.94 & 66.72  & 71.49     & 88.90  & 4.12 & 75.01  & 76.91     & 85.22 & \underline{4.32} & 77.61  & 76.54     \\
ExTrans-7B                & \textbf{90.55}$^{\dagger}$ & \textbf{4.60}$^{\dagger}$  & \textbf{82.29}$^{\dagger}$  & \textbf{74.23}$^{\dagger}$     & \textbf{90.86}$^{\dagger}$ & \textbf{4.25}$^{\ddagger}$ & \textbf{79.24}$^{\dagger}$  & \underline{78.02}$^{\dagger}$     & 86.15 & \textbf{4.35}$^{\dagger}$ & \textbf{80.34}$^{\dagger}$  & 77.67$^{\dagger}$    \\ \bottomrule[1pt]
\end{tabular}
}
\caption{Experimental results in English-to-Chinese literary translation. The \textbf{bold} and the \underline{underline} denote the best and second-best scores, respectively. ``$\dagger$'' and``$\ddagger$'' denote statistically significant better than the DeepTrans-7B~\cite{wang2025deep} with t-test p < 0.01 and 0.05, respectively. ``$\diamondsuit$'' denotes models are trained on MetaphorTrans.}
\label{table:main_res_in_en2zh}
\end{table*}

\subsection{Baselines}

\noindent \textbf{Non-reasoning LLMs.}
We leverage Llama-3.1-8B-Instruct~\cite{grattafiori2024llama}, Qwen2.5-7B-Instruct, Qwen2.5-14B-Instruct~\cite{yang2024qwen2} and GPT-4o~\cite{hurst2024gpt} as baselines.
We also fine-tune LLama-3.1-8B-Instruct, Qwen2.5-7B-Instruct and Qwen2.5-14B-Instruct with only paired sentences of the MetaphorTrans training data (without long CoT).
The fine-tuned LLMs are denoted as Llama-3.1-8B-MT, Qwen2.5-7B-MT and Qwen2.5-14B-MT.

\vspace{0.5ex}
\noindent \textbf{LRMs.}
QwQ-32B(-preview)~\cite{team2024qwq}, Marco-o1-7B~\cite{zhao2024marco}, DeepSeek-Qwen-7B, DeepSeek-Llama-8B, DeepSeek-Qwen-14B, DeepSeek-Qwen-32B, DeepSeek-R1~\cite{guo2025deepseek} and o1-preview~\cite{openai_o1_2024} are used as baselines.
More recently, Qwen3 LLM family\footnote{\url{https://github.com/QwenLM/Qwen3}} is proposed, and we use Qwen3-8B, Qwen3-14B and Qwen3-32B as baselines.
The Qwen3 LLMs support both reasoning and non-reasoning modes, which we denote as Qwen3 (w/ CoT) and Qwen3 (w/o CoT), respectively.
We also use previous MT LRMs as our baselines, including DRT-7B, DRT-8B, DRT-14B~\cite{wang2024drt} and DeepTrans-7B~\cite{wang2025deep}.

\subsection{Results \& Analyses}

Table~\ref{table:main_res_in_en2zh} shows the experimental results in English-to-Chinese literary translation.

\vspace{0.5ex}
\noindent \textbf{Compare with previous MT LRMs.}
ExTrans-7B significantly outperforms previous MT LMRs, \emph{i.e.}, DRT-7B/8B/14B~\cite{wang2024drt} and DeepTrans-7B~\cite{wang2025deep}.
Specifically, on the MetaphorTrans test set, ExTrans-7B outperforms DRT-7B by 5.8\%, 13.6\%, 9.6\% and 3.4\% in terms of GRF, GEA5, GEA100 and CometKiwi, respectively.
Compared with DeepTrans-7B, the counterpart improvements are 1.9\%, 9.3\%, 9.2\% and 3.4\%, showing the effectiveness of our reward modeling method.
With the help of the strong exemplar, ExTrans-7B can effectively enhance its MT capabilities by comparing itself with the exemplar.

\vspace{0.5ex}
\noindent \textbf{Compare with baselines that are trained on MetaphorTrans.}
Similar to ExTrans-7B, some baselines are also trained on MetaphorTrans, which are marked by ``$\diamondsuit$'' in Table~\ref{table:main_res_in_en2zh}. We find that ExTrans-7B outperforms them not only on the MetaphorTrans test set, but also on the other two literary books.
This finding also verifies the effectiveness of our reward modeling.
The improvement brought by RL can be generalized to other genres.

\vspace{0.5ex}
\noindent \textbf{Compare with strong LRM baselines.}
We find that ExTrans-7B outperforms OpenAI-o1 and DeepSeek-R1 on MetaphorTrans, and achieves the new state-of-the-art performance in terms of all metrics, showing its superiority.
In \emph{O. Henry} and \emph{Orbital}, ExTrans-7B also achieves the best performance in terms of most metrics.
Comparing ExTrans-7B with its cold start version, \emph{i.e.}, ExTrans-7B (cold start), the significant improvement brought by the RL training is demonstrated.

\begin{figure*}[t]
\centering
\subfigure[GRF and GEA100]{
  \includegraphics[width=0.23\linewidth]{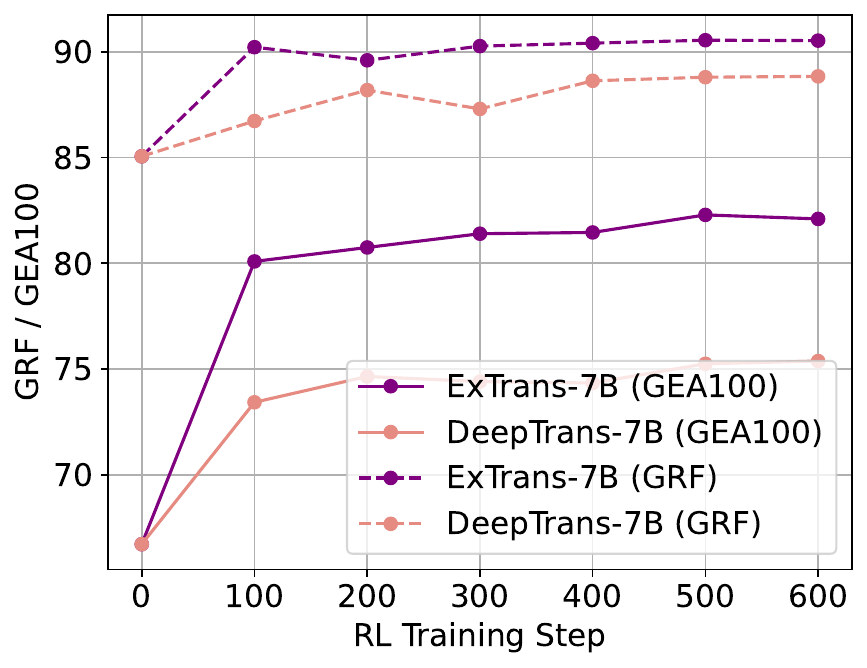}
}
\subfigure[CometKiwi]{
  \includegraphics[width=0.23\linewidth]{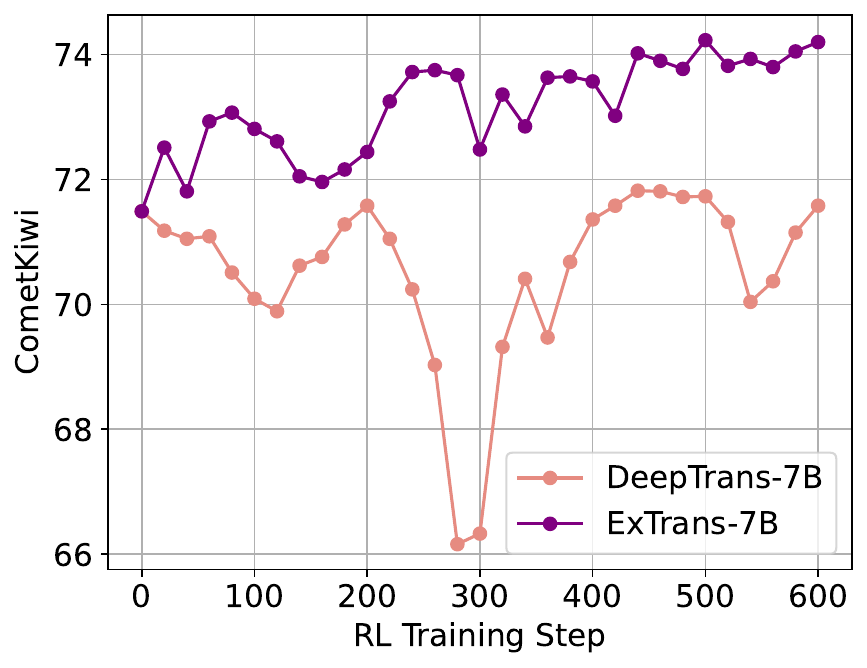}
}
\subfigure[$r_\text{all}$]{
  \includegraphics[width=0.23\linewidth]{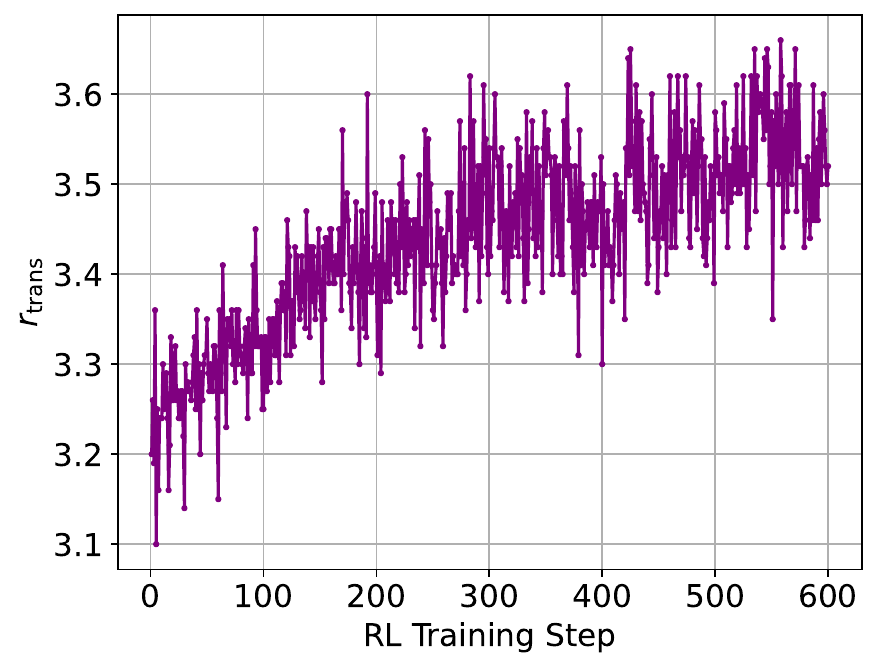}
}
\subfigure[$r_\text{cometk}$]{
  \includegraphics[width=0.23\linewidth]{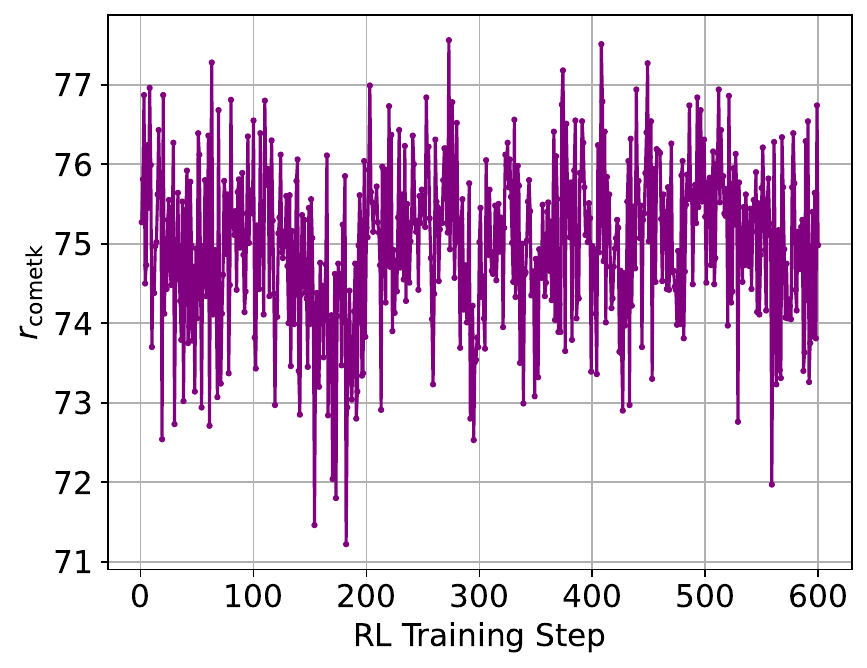}
}
\caption{Performance and Rewards change of ExTrans-7B during RL training. Among them, (a) and (b) are conducted on the MetaphorTrans test set, while (c) and (d) are conducted on its training set. The horizontal axis denotes the number of training steps, and there are 600 steps (2 epochs) in total. The vertical axis denotes the value of the corresponding metrics.}
\label{fig:intermediate_analyses}
\end{figure*}

\vspace{0.5ex}
\noindent \textbf{Intermediate-Stage Analyses.}
To provide a deeper understanding of ExTrans-7B, we analyze the model performance and reward change during the RL training stage.
Figure~\ref{fig:intermediate_analyses} shows the details, and we analyze it from the following aspects:

In terms of \emph{GPT-based evaluation metrics}, we compare the changes between ExTrans-7B and previous DeepTrans-7B~\cite{wang2025deep}.
As shown in Figure~\ref{fig:intermediate_analyses} (a), the performance of ExTrans-7B is consistently better than that of DeepTrans-7B.
After only 100 steps of RL training, ExTrans-7B is able to beat the final DeepTrans-7B.

In terms of \emph{CometKiwi}, ExTrans-7B also show its superiority compared with DeepTrans-7B (c.f. Figure~\ref{fig:intermediate_analyses} (b)).
The CometKiwi score of ExTrans-7B generally increases along with the training process.

In terms of \emph{reward values}, we find that the exemplar-enhanced translation reward (\emph{i.e.}, $r_\text{trans}$) generally increases during RL training, while the CometKiwi reward (\emph{i.e.}, $r_\text{cometk}$) continues to fluctuate.
Both rewards are important for training our ExTrans-7B, which we will further discuss in \S~\ref{subsec:ablations}.

\begin{table}[t]
\centering
\resizebox{0.45\textwidth}{!}
{
\begin{tabular}{lcccc}
\toprule[1pt]
\multicolumn{1}{c}{\multirow{2}{*}{Model}} & GRF   & GEA5 & GEA100 & CometKiwi \\
\multicolumn{1}{c}{}                       & \multicolumn{4}{c}{MetaphorTrans} \\ \midrule[1pt]
ExTrans-7B                                 & 90.55 & 4.60  & 82.29  & 74.23     \\
ExTrans-7B (w/o $r_\text{trans}$)                     & 88.10  & 4.34 & 76.90   & 76.79     \\
ExTrans-7B (w/o $r_\text{cometk}$)                    & 90.04 & 4.62 & 82.39  & \color{red}{60.32}     \\\midrule[1pt]
                                           & \multicolumn{4}{c}{\emph{O. Henry}}      \\\midrule[1pt]
ExTrans-7B                                 & 90.86 & 4.25 & 79.24  & 78.02     \\
ExTrans-7B (w/o $r_\text{trans}$)                     & 88.61 & 4.08 & 74.65  & 79.79     \\
ExTrans-7B (w/o $r_\text{cometk}$)                    & 89.14 & 4.54 & 78.86  & \color{red}{69.78}    \\ \bottomrule[1pt]
\end{tabular}
}
\caption{The experimental results of ablation studies.}
\label{table:ablations}
\end{table}

\vspace{0.5ex}
\noindent \textbf{Human Evaluation.}
To verify the superiority of ExTrans-7B, we also employ human evaluation on the translations of ExTrans-7B, DeepTrans-7B, QwQ-32B and Qwen3-32B (w/ CoT).
The results are provided in Appendix~\ref{appendix:human_evaluation}.

\subsection{Ablation Study}
\label{subsec:ablations}

There are four types of rewards in $r_\text{all}$, \emph{i.e.}, $r_\text{format}$, $r_\text{thought}$, $r_\text{trans}$ and $r_\text{cometk}$ (c.f. Eq.~\ref{eq:reward_in_single_direction}).
Among them, $r_\text{format}$ and $r_\text{thought}$ are basic rewards and have already been verified in \citet{wang2025deep}.
We further verify the effectiveness of the exemplar-enhanced translation reward ($r_\text{trans}$) and cometkiwi reward ($r_\text{cometk}$).
To this end, we design the following two variants of ExTrans-7B:
(1) ExTrans-7B (w/o $r_\text{trans}$) removes $r_\text{trans}$ in the reward modeling during RL training; while (2) ExTrans-7B (w/o $r_\text{cometk}$) removes $r_\text{cometk}$.
As shown in Table~\ref{table:ablations}, when removing the exemplar-enhanced translation reward, the model performance in terms of GPT-4o metrics decreases.
When removing the CometKiwi reward, the CometKiwi score will significantly decrease, \emph{e.g.}, 74.23$\rightarrow$60.32.
These findings validate the rationality behind our reward modeling design.

\begin{table*}[t]
\centering
\resizebox{0.98\textwidth}{!}
{
\begin{tabular}{l|cc|cc|cc|cc}
\toprule[1pt]
\multicolumn{1}{c|}{Model} & \diagbox[dir=NW]{Trg}{Src}           & English (MetaphorTrans)                      & \diagbox[dir=NW]{Trg}{Src}           & English (MetaphorTrans)                     & \diagbox[dir=NW]{Trg}{Src}           & Russian (Par3)                     & \diagbox[dir=NW]{Trg}{Src}           & Russian (Par3)                     \\ \midrule[1pt]
Qwen2.5-7B-Instruct       & \multirow{5}{*}{Ar} & 54.20 / 2.53 / 34.00 / 57.49 & \multirow{5}{*}{It} & 79.11 / 3.50 / 59.90 / 69.14 & \multirow{5}{*}{Ar} & 62.62 / 2.96 / 44.20 / 59.09 & \multirow{5}{*}{Fr} & 82.25 / 3.83 / 66.44 / 68.19 \\
QwQ-32B                   &                     & 78.29 / 3.82 / 60.60 / 63.55 &                     & 86.64 / 3.95 / 73.97 / 72.20 &                     & 81.44 / 3.96 / 67.60 / 62.47 &                     & 88.27 / 4.17 / 77.60 / 69.26 \\
o1-preview                &                     & 90.18 / 4.20 / 76.43 / 67.75 &                     & 91.02 / 4.10 / 80.60 / 75.53 &                     & 75.52 / 3.53 / 55.52 / 62.86 &                     & 92.69 / 4.20 / 81.90 / 71.24 \\
mExTrans-7B (cold start)   &                     & 72.05 / 3.43 / 48.10 / 63.67 &                     & 80.25 / 3.62 / 62.40 / 70.67 &                     & 71.00 / 3.39 / 53.90 / 62.07 &                     & 83.71 / 3.96 / 68.83 / 68.87 \\
mExTrans-7B                &                     & 81.10 / 4.01 / 61.26 / 67.22 &                     & 83.94 / 3.93 / 67.85 / 73.21 &                     & 80.06 / 3.76 / 61.58 / 65.13 &                     & 86.56 / 4.13 / 73.88 / 68.77 \\ \midrule[1pt]
Qwen2.5-7B-Instruct       & \multirow{5}{*}{Cs} & 54.92 / 2.30 / 29.55 / 56.92 & \multirow{5}{*}{Ja} & 69.73 / 3.16 / 48.23 / 69.22 & \multirow{5}{*}{Cs} & 64.11 / 2.83 / 41.08 / 66.11 & \multirow{5}{*}{It} & 82.36 / 3.79 / 65.52 / 70.27 \\
QwQ-32B                   &                     & 75.69 / 3.47 / 55.18 / 64.67 &                     & 84.72 / 4.17 / 71.30 / 75.36 &                     & 81.33 / 3.71 / 63.17 / 71.18 &                     & 89.28 / 4.22 / 77.05 / 71.47 \\
o1-preview                &                     & 89.02 / 4.16 / 77.71 / 73.26 &                     & 87.64 / 4.46 / 79.45 / 78.52 &                     & 92.77 / 4.51 / 81.47 / 77.58 &                     & 92.56 / 4.38 / 81.97 / 73.79 \\
mExTrans-7B (cold start)   &                     & 59.20 / 2.38 / 32.76 / 59.21 &                     & 80.70 / 3.91 / 61.66 / 74.32 &                     & 65.17 / 2.90 / 43.75 / 67.80 &                     & 82.06 / 3.86 / 66.95 / 70.71 \\
mExTrans-7B                &                     & 66.52 / 2.78 / 37.84 / 63.12 &                     & 84.89 / 4.29 / 71.17 / 76.88 &                     & 71.92 / 3.27 / 49.75 / 70.47 &                     & 85.28 / 4.13 / 72.10 / 72.05 \\ \midrule[1pt]
Qwen2.5-7B-Instruct       & \multirow{5}{*}{De} & 76.95 / 3.25 / 52.70 / 68.02 & \multirow{5}{*}{Ko} & 56.83 / 2.61 / 37.82 / 64.39 & \multirow{5}{*}{De} & 74.41 / 3.43 / 57.02 / 66.91 & \multirow{5}{*}{Ja} & 74.30 / 3.36 / 53.20 / 63.84 \\
QwQ-32B                   &                     & 83.21 / 3.79 / 70.62 / 70.04 &                     & 81.42 / 3.99 / 67.84 / 71.90 &                     & 84.20 / 3.94 / 73.17 / 69.29 &                     & 83.67 / 4.29 / 72.04 / 67.76 \\
o1-preview                &                     & 89.12 / 4.04 / 79.92 / 74.02 &                     & 88.78 / 4.55 / 80.62 / 77.82 &                     & 91.53 / 4.21 / 81.97 / 72.63 &                     & 90.83 / 4.61 / 82.36 / 71.77 \\
mExTrans-7B (cold start)   &                     & 78.41 / 3.46 / 56.00 / 68.91 &                     & 77.67 / 3.51 / 52.42 / 71.94 &                     & 81.58 / 3.77 / 64.22 / 68.70 &                     & 82.47 / 3.90 / 65.23 / 68.45 \\
mExTrans-7B                &                     & 82.98 / 3.84 / 65.22 / 71.60 &                     & 83.08 / 4.09 / 65.28 / 74.95 &                     & 84.98 / 4.01 / 69.67 / 69.88 &                     & 85.08 / 4.21 / 71.26 / 69.88 \\ \midrule[1pt]
Qwen2.5-7B-Instruct       & \multirow{5}{*}{Es} & 83.03 / 3.69 / 64.58 / 68.79 & \multirow{5}{*}{Ru} & 67.45 / 3.05 / 47.12 / 63.98 & \multirow{5}{*}{En} & 83.05 / 4.07 / 73.02 / 75.95 & \multirow{5}{*}{Ko} & 63.98 / 2.78 / 40.32 / 60.57 \\
QwQ-32B                   &                     & 89.14 / 4.00 / 76.22 / 72.29 &                     & 85.50 / 3.99 / 70.78 / 69.87 &                     & 86.17 / 4.14 / 78.03 / 75.94 &                     & 78.14 / 3.92 / 65.95 / 64.25 \\
o1-preview                &                     & 92.22 / 4.12 / 82.66 / 73.90 &                     & 90.30 / 4.19 / 77.67 / 73.30 &                     & 90.92 / 4.32 / 84.00 / 77.24 &                     & 90.47 / 4.66 / 82.97 / 70.51 \\
mExTrans-7B (cold start)   &                     & 85.65 / 3.98 / 70.38 / 71.08 &                     & 79.78 / 3.65 / 57.30 / 67.94 &                     & 84.30 / 4.00 / 72.60 / 75.78 &                     & 79.89 / 3.71 / 59.15 / 65.17 \\
mExTrans-7B                &                     & 87.05 / 4.17 / 75.08 / 73.00 &                     & 84.26 / 4.05 / 66.22 / 71.68 &                     & 86.50 / 4.26 / 77.40 / 75.99 &                     & 83.98 / 4.01 / 66.08 / 67.08 \\ \midrule[1pt]
Qwen2.5-7B-Instruct       & \multirow{5}{*}{Fr} & 80.54 / 3.58 / 59.23 / 69.43 & \multirow{5}{*}{Zh} & 81.53 / 3.62 / 66.21 / 70.36 & \multirow{5}{*}{Es} & 78.06 / 3.79 / 65.14 / 67.99 & \multirow{5}{*}{Zh} & 84.13 / 3.89 / 69.22 / 65.80 \\
QwQ-32B                   &                     & 89.39 / 4.02 / 75.92 / 73.01 &                     & 88.06 / 4.09 / 74.38 / 72.88 &                     & 88.61 / 4.26 / 78.55 / 70.89 &                     & 89.15 / 4.32 / 77.70 / 66.82 \\
o1-preview                &                     & 91.92 / 4.09 / 78.90 / 75.14 &                     & 87.11 / 4.06 / 78.01 / 73.70 &                     & 92.48 / 4.38 / 82.05 / 72.16 &                     & 91.78 / 4.44 / 82.12 / 68.81 \\
mExTrans-7B (cold start)   &                     & 84.17 / 3.83 / 64.90 / 71.45 &                     & 86.17 / 4.36 / 72.99 / 71.77 &                     & 83.10 / 4.02 / 66.75 / 69.24 &                     & 87.69 / 4.17 / 73.20 / 66.85 \\
mExTrans-7B                &                     & 87.31 / 4.03 / 72.95 / 73.68 &                     & 90.44 / 4.54 / 80.36 / 74.16 &                     & 86.24 / 4.21 / 72.58 / 70.03 &                     & 90.32 / 4.45 / 79.70 / 67.13 \\ \bottomrule[1pt]
\end{tabular}
}
\caption{Experimental results in multilingual literary translation (in terms of GRF / GEA5 / GEA100 / CometKiwi). ``Src'' and ``Trg'' denote the source and the target language, respectively. The English sources are selected from the MetaphorTrans test set~\cite{wang2024drt}, while the Russian sources are selected from Par3~\cite{thai-etal-2022-exploring}.}
\label{table:main_res_in_multilingual}
\end{table*}

\subsection{Multilingual Generalization}

As described in \S~\ref{subsec:2.2}, we extend the reward modeling to the multilingual settings via a lightweight method, and train mExTrans-7B LRM.
Table~\ref{table:main_res_in_multilingual} shows the experimental results in English-to-X and Russian-to-X literary translation.
We compare mExTrans-7B with its backbone, \emph{i.e.}, Qwen2.5-7B-Instruct, and its cold-start version, \emph{i.e.}, mExTrans-7B (cold start). Besides, QwQ-32B and o1-preview serve as strong LRM baselines for comparison.
We analyze the results from the following aspects:

\vspace{0.5ex}
\noindent \textbf{Qwen2.5-7B-Instruct $\rightarrow$ mExTrans-7B (cold start):}
The cold start SFT (c.f. \S~\ref{subsec:2.3}) significantly improves the model performance in most translation directions, showing the effectiveness of powerful data created by the advanced LRMs~\cite{guo2025deepseek,yang2025qwen3}.
For example, in En$\rightarrow$Ar translation, mExTrans-7B (cold start) outperforms Qwen2.5-7B-Instruct by 17.85 GRF, 0.90 GEA5, 14.10 GEA100, and 6.18 CometKiwi; while the counterpart improvements in Ru$\rightarrow$Ar are 8.38 GRF, 0.43 GEA5, 9.70 GEA100 and 2.98 CometKiwi.

\vspace{0.5ex}
\noindent \textbf{mExTrans-7B (cold start) $\rightarrow$ mExTrans-7B:}
With the help of the designed reward modeling method, mExTrans-7B further improves the model performance in the multilingual literary translation.
We find that mExTrans-7B outperforms mExTrans-7B (cold start) in almost all directions in terms of all metrics, demonstrating the effectiveness and efficiency of the lightweight reward modeling method.
By solely verifying the correctness of the generation format and target language in other directions, the method can transfer MT capabilities from a single high-resource direction (\emph{i.e.}, English-to-Chinese in our work) to others.

\vspace{0.5ex}
\noindent \textbf{Compare with strong LRM baselines:}
In terms of CometKiwi, we find that mExTrans-7B outperforms QwQ-32B in most directions, while QwQ-32B performs better in terms of GPT-4o evaluation.
Compared with QwQ-32B, the performance of mExTrans-7B is competitive.
However, when comparing mExTrans-7B with o1-preview, we observe that o1-preview consistently outperforms mExTrans-7B across most metrics, indicating that a performance gap remains when transferring MT capabilities to multilingual settings.

\section{Related Work}

In recent years, large reasoning models (LRMs), \emph{e.g.}, OpenAI o1~\cite{openai_o1_2024} and DeepSeek-R1~\cite{guo2025deepseek}, have pioneered a growing research in long chain-of-thought (CoT) reasoning.
Many studies make their efforts, and bring the success of LRMs to different tasks~\cite{chen2025towards,li2025system,zhang2024o1,guan2025deeprag,jin2025search,li2025search}.
Some researchers investigate the MT capability of deep reasoning LLMs.
\citet{zhao2024marco} and \citet{liu2025new} discuss the potential of long CoT reasoning in MT with some heuristic examples.
\citet{wang2024drt} argue that literary sentences that involve metaphors or similes are hard to translate, and might be suitable for LRMs.
Based on this motivation, they construct MetaphorTrans dataset, and further train DRT LRMs by supervised fine-tuning on MetaphorTrans.
R1-T1~\cite{he2025r1} and MT-R1~\cite{feng2025mt} leverage Comet, CometKiwi or BLEU as reward signals to train LRMs.
In view of the strong ability of LLM-as-a-judge, DeepTrans-7B~\cite{wang2025deep} uses DeepSeek-v3 to provide reference-free evaluation scores during RL training.
Different from the above studies, we explore a new reward modeling strategy to leverage the strong abilities of both LLM-as-a-judge and LLM-as-an-exemplar. Besides, we also generalize the MT LRM to multilingual settings via a lightweight method.

\section{Conclusion}

In this paper, we aim to improve the MT ability of LRMs via RL.
In detail, we propose a new reward modeling method that employs a strong LRM (\emph{i.e.}, DeepSeek-R1) as an exemplar, and compares the policy model with the exemplar to provide reward signals. In this way, we train ExTrans-7B LRM via RL.
In addition, we design a lightweight method to extend ExTrans-7B to the multilingual settings with 11 languages, resulting in mExTrans-7B.
Experimental results in literary translation demonstrate the effectiveness and superiority of the proposed methods.
ExTrans-7B achieves the state-of-the-art performance, and it outperforms previous MT LRMs and general LRMs by a large margin.

\section*{Limitations}

While we show the effectiveness of ExTrans-7B and mExTrans-7B, there are some limitations worth noting:
(1) ExTrans-7B needs exemplar translations during RL training, which might be costly. For each training sample, we need to infer an advanced LRM (\emph{e.g.}, DeepSeek-R1-671B) to obtain the translations.
(2) Though mExTrans-7B generalizes its multilingual ability, it still has a gap behind the advanced LRMs (\emph{e.g.}, o1-preview) in the low-resource or unrepresentative translation directions. Future work could explore more effective methods to generalize the multilingual MT abilities of LRMs.

\section*{Ethical Considerations}

We discuss the main ethical considerations of (m)ExTrans-7B as follows:
(1) \emph{Licenses.} We will release our model checkpoints under CC-BY-NC-SA 4.0 license.
(2) \emph{Toxicity.} The backbone of (m)ExTrans-7B is Qwen2.5-7B-Instruct. Besides, during RL training, we use DeepSeek-v3 as the reward model to score the translation quality. Therefore, (m)ExTrans-7B might involve the same biases and toxic behaviors exhibited by these LLMs.

\bibliography{custom}


\appendix

\section{Thought Reward Prompt}
\label{appendix:thought_reward_prompt}

We borrow the prompt from \citet{wang2025deep}:

\begin{tcolorbox}
\fontsize{10pt}{11pt}\selectfont

A translation question requires translating a given text from [src lang] into [trg lang]. \\

The given text is as follows:\\
<text>\\
\{src\}\\
</text>\\

Someone did this translation question, and began to think how to translate:\\
<think>\\
\{think\}\\
</think>\\

Please judge whether there is a detailed analysis of the given text in this thinking process:\\
1. No analysis: Only very shallow thinking was done, and no detailed analysis of the given text was carried out.\\
2. Slight analysis: The given text was analyzed in detail, and how to translate it was discussed in detail.\\
3. Detailed analysis: The given text was analyzed in detail, and various translation possibilities were discussed in detail, and trade-offs were made.\\

Please directly output your judgment results, such as: ``no analysis'', ``slight analysis'' or ``detailed analysis''

\end{tcolorbox}

\section{Evaluation Prompt}
\label{appendix:evaluation_prompt}

The evaluation prompt of GRF borrows from \citet{kocmi-federmann-2023-large}, and is also employed in \citet{wang2024drt,wang2025deep}:

\begin{tcolorbox}
\fontsize{10pt}{11pt}\selectfont

Score the following translation from \texttt{[src lang]} to \texttt{[trg lang]} on a continuous scale from 0 to 100, where score of zero means "no meaning preserved" and score of one hundred means "perfect preservation of meaning, with faithfulness, expressiveness, and elegance".\\

\texttt{[src lang]} source: \{src\}\\
\texttt{[trg lang]} translation: \{hyp\}\\

Score:

\end{tcolorbox}

The prompt evaluates a translation from a general perspective, and achieves a high correlation with humans~\cite{kocmi-federmann-2023-large}.

The GEA100 prompt borrows from \citet{wang2024drt}, which includes a system prompt and a user prompt:

\begin{tcolorbox}
\fontsize{10pt}{11pt}\selectfont

\texttt{SYSTEM PROMPT:} \\

Please evaluate the following Chinese translation of an English text. Rate the translation on a scale of 0 to 100, where:\\
- 10 points: Poor translation; the text is somewhat understandable but contains significant errors and awkward phrasing that greatly hinder comprehension for a Chinese reader.\\
- 30 points: Fair translation; the text conveys the basic meaning but lacks fluency and contains several awkward phrases or inaccuracies, making it challenging for a Chinese reader to fully grasp the intended message.\\
- 50 points: Good translation; the text is mostly fluent and conveys the original meaning well, but may have minor awkwardness or slight inaccuracies that could confuse a Chinese reader.\\
- 70 points: Very good translation; the text is smooth and natural, effectively conveying the intended meaning, but may still have minor issues that could slightly affect understanding for a Chinese reader.\\
- 90 points: Excellent translation; the text is fluent and natural, conveying the original meaning clearly and effectively, with no significant issues that would hinder understanding for a Chinese reader.\\

Please provide the reason first, followed by a score. Format your evaluation in the JSON structure below:\\
\{"reason": "reason for the score", "score": int\}\\

\end{tcolorbox}

\begin{tcolorbox}
\fontsize{10pt}{11pt}\selectfont
\texttt{USER PROMPT:} \\

<text>\\
\{src\}\\
</text>\\
<translation>\\
\{trans\}\\
</translation>
\end{tcolorbox}

The GEA100 prompt evaluates a translation from a literary perspective.
The GEA5 prompt simply narrows the scoring scope of GEA100 from a 100-point to a 5-point scale.

\section{Implementation Details.}
\label{appendix:implementation_details}

\noindent \textbf{Cold Start SFT.}
Llama-Factory framework~\cite{zheng-etal-2024-llamafactory} is used during the SFT stage.
We conduct experiments on 8$\times$NVIDIA H20 GPUs (96G) with 1e-5 learning rate and 8 (8$\times$1 accumulation step) batch size.
DeepSpeed ZeRO-3 optimization~\cite{rasley2020deepspeed} is also used during SFT.
We set the number of SFT epochs to 2, and it costs about 1 GPU hour for ExTrans-7B while 3 GPU hours for mExTrans-7B.

To create the cold-start SFT data, DeepSeek-R1 (671B) is deployed on 2$\times$NVIDIA H20 GPUs (96G).
During generation, we set the temperature to 0.1, and the top-p to 0.95.

\vspace{0.5ex}
\noindent \textbf{RL Training.}
We use GRPO RL algorithm implemented by verl\footnote{\url{https://github.com/volcengine/verl}}~\cite{sheng2024hybridflow}.
2$\times$8 H20 GPUs are used, where 8 GPUs are used to deploy DeepSeek-v3 (awq quantization) as the reward model, and another 8 GPUs are used to optimize the policy model.
We set the batch size to 64, the learning rate to 1e-6, the rollout number to 8 and the rollout temperature to 0.6, and the KL loss coefficient to 1e-3.
The number of training epochs is set to 2.
The hyperparameter $\alpha$ and $\beta$ in Eq.~\ref{eq:reward_in_single_direction} are both set to 1.0.
In this way, the CometKiwi reward $r_\text{cometk}$ retains auxiliary status compared with the exemplar-enhanced translation reward $r_\text{trans}$.
The RL training costs 1.0K GPU hours for ExTrans-7B, while 1.2K GPU hours for mExTrans-7B.

\vspace{0.5ex}
\noindent \textbf{Evaluation.}
When evaluating model performance on the test set, we use vLLM toolkit\footnote{\url{https://github.com/vllm-project/vllm}} to accelerate the model generation.
We use the sampling decoding strategy with 0.1 temperature.

To calculate CometKiwi, we leverage the official codes\footnote{\url{https://github.com/Unbabel/COMET}} and the official models\footnote{\url{https://huggingface.co/Unbabel/wmt22-cometkiwi-da}}.
When calculating the GPT-4o evaluation metrics (GRF, GEA100 and GEA5), we set the temperature to 0.1.

Since GPT-4o needs API costs, we randomly select 400 samples from each test set (MetaphorTrans, \emph{O. Henry} and \emph{Orbital}) to evaluate ExTrans-7B.
To evaluate mExTrans-7B in the multilingual settings, in En$\rightarrow$Zh, we randomly select 400 samples; while in other directions, we randomly select 200 samples to conduct GPT-4o evaluation.

\begin{table}[t]
\centering
\resizebox{0.38\textwidth}{!}
{
\begin{tabular}{lccc}
\toprule[1pt]
\multicolumn{1}{c}{Model} & Flu.   & Sem.   & Lit.   \\ \midrule[1pt]
QwQ-32B      & -0.135 & -0.160 & -0.155 \\
Qwen3-32B (w/ CoT)          & -0.025 & -0.055  &  -0.130 \\
DeepTrans-7B           & 0.035  & 0.085  & 0.090  \\
ExTrans-7B                   & \textbf{0.125}  & \textbf{0.130}  & \textbf{0.195}  \\ \bottomrule[1pt]
\end{tabular}
}
\caption{Human evaluation results (Flu.: fluency; Sem.: semantic accuracy; Lit.: literary quality).}
\label{table:human_eval}
\end{table}

\section{Human Evaluation}
\label{appendix:human_evaluation}

We conduct human evaluation to further evaluate the performance of ExTrans-7B.
Following~\citet{wang2024drt}, we randomly select 200 samples from the MetaphorTrans test set, and employ three human evaluators with high levels of fluency in English and Chinese to assess the generated translations from three aspects: fluency (Flu.), semantic accuracy (Sem.) and literary quality (Lit.).
Following the Best-Worst Scaling method~\cite{kiritchenko-mohammad-2017-best}, evaluators are asked to select the best and the worst generated translation on each aspect.
The result scores are calculated based on the percentage of times each model is selected as best minus the times it is selected as worst. Thus, the final scores should range from -1 (worst) to 1 (best).
Table~\ref{table:human_eval} shows the results of human evaluation, and we can find that ExTrans-7B outperforms other LRMs, demonstrating its effectiveness.
The Fleiss’ Kappa scores~\cite{fleiss1971measuring} of Flu., Sem. and
Lit. are 0.81, 0.63 and 0.70, respectively, indicating a good inter-agreement among evaluators.

\end{document}